\documentclass[sensors,article,accept,pdftex,moreauthors]{Definitions/mdpi} 
\firstpage{1} 
\pubvolume{26}
\issuenum{16}
\articlenumber{5046}
\pubyear{2026}
\copyrightyear{2026}
\externaleditor{Tie Zhang} % More than 1 editor, please add `` and '' before the last editor name
\datereceived{5 June 2026} 
\daterevised{3 August 2026} % Comment out if no revised date
\dateaccepted{4 August 2026} 
\datepublished{8 August 2026} 
\Title{Action- and Language-Conditioned Video Assessment for Embodied Control}

\Author{Hwanhee Kim
 $^{1}$\orcidA{}, Jaehyun Jang $^{2}$\orcidB{}, Seungmin Cha $^{2}$, Hyeonseo Yun $^{2}$, Donghoon Lee $^{2}$ and Chang D. Yoo $^{2,}$*}

\AuthorNames{Hwanhee Kim, Jaehyun Jang, Seungmin Cha, Hyeonseo Yun, Donghoon Lee and Chang D. Yoo}

\address{%
$^{1}$ \quad Robotics Program, Korea Advanced Institute of Science and Technology (KAIST), Daejeon 34141, Republic~of~Korea; khhandrea@kaist.ac.kr\\
$^{2}$ \quad School of Electrical Engineering, Korea Advanced Institute of Science and Technology (KAIST), Daejeon~34141, Republic of Korea; jhjangjh@kaist.ac.kr (J.J.); cseungmin7@kaist.ac.kr (S.C.); hyunseo@kaist.ac.kr (H.Y.); dh\_lee99@kaist.ac.kr (D.L.)}

\corres{Correspondence: cd\_yoo@kaist.ac.kr (C.D.Y.)}

\abstract{Vision-based embodied agents executing multi-step natural language instructions require feedback mechanisms that assess task progress over complete trajectories. Conventional approaches based on final-frame matching or continuous embedding similarity may overlook intermediate transitions that are necessary for determining whether an instruction has been completed. We propose \ours{} (\fullours), a trajectory evaluator that conditions its assessment on visual observations, the executed action sequence, and the natural language instruction. The method uses a pre-trained vision-language model (VLM) in two stages: it first summarizes frame-to-frame visual transitions conditioned on the executed actions and then assesses the generated summary with respect to the instruction to produce a discrete trajectory-level progress score. In simulated 3D household environments, \ours{} exhibits a conservative assessment pattern with near-zero false-positive rates. When used as terminal feedback for closed-loop policy optimization, it provides more effective feedback than the evaluated static image and embedding-based visual baselines and reduces the performance gap to a ground-truth oracle. These results support action- and language-conditioned video assessment as an interpretable feedback mechanism for the evaluated simulated embodied-control tasks.}

\keyword{vision-language models; video assessment; trajectory evaluation; embodied control; task-progress feedback; embodied AI}

\newcommand{\ours}{ALVA}
\newcommand{\fullours}{Action- and Language-Conditioned Video Assessment}

\newcommand{\stdv}[1]{\scalebox{0.7}{$\pm$#1}}

\usepackage{algorithm}
\usepackage{algorithmic}
\usepackage{subcaption}
\usepackage{arxiv-neutral} % arXiv-only presentation layer; manuscript content unchanged
\begin{document}

%%%%%%%%%%%%%%%%%%%%%%%%%%%%%%%%%%%%%%%%%%
\section{Introduction} \label{sec:introduction}

Embodied agents increasingly rely on visual observations to execute sequential actions in complex environments. For~multi-step tasks specified by natural language, reliable policy optimization requires an evaluator that can assess how a trajectory progresses toward the instruction rather than considering only an isolated final observation~\cite{tellex2011understanding, anderson2018vision, shridhar2020alfred}.

A common zero-shot strategy compares a final visual observation with a language instruction using pre-trained vision-language representations~\cite{radford2021learning, mahmoudieh2022zero, rocamonde2024vision}.~This approach is simple, but~it can discard intermediate task progress and cannot directly indicate which executed actions correspond to the observed visual changes. Video-language and robotic representation learning methods incorporate temporal information~\cite{nair2022r3m, ma2023liv, sontakke2024roboclip}, yet many still derive feedback from embedding similarity rather than an explicit interpretation of action-conditioned~transitions.

We therefore formulate trajectory assessment as an action- and language-conditioned video evaluation problem. The~evaluator receives a video trajectory, the~executed action sequence, and~a natural language instruction, and~determines whether the observed transitions indicate meaningful task progress. This formulation focuses on action-to-transition correspondence and task-progress~assessment.

To address this problem, we propose \ours{} (\fullours). The~system represents each interaction as a video trajectory composed of visual observations, executed control commands, and~a natural language instruction. A~pre-trained vision-language model~\cite{openai2023gpt4v, reid2024gemini, openai2024gpt4o} is used as a trajectory-level evaluator through a two-stage querying process. In~Stage 1, the~model performs action-conditioned transition summarization by describing frame-to-frame visual changes in relation to the executed action sequence. In~Stage 2, the~generated summary is assessed with respect to the language instruction to estimate task progress. The~resulting discrete score provides trajectory-level feedback without relying only on static final-frame~matching.

We evaluate \ours{} in simulated 3D household environments based on ALFRED and AI2-THOR, where agents execute sequential actions to complete natural language instructions~\cite{shridhar2020alfred, kolve2017ai2}. The~results show that \ours{} provides more effective feedback than the evaluated static image-text and embedding-based visual baselines~\cite{radford2021learning, nair2022r3m, sontakke2024roboclip}. The~findings are limited to the tested simulated household tasks and VLM~backbones.

The main contributions of this study are as follows:

\begin{itemize}
    \item We propose \ours{}, an~action- and language-conditioned video assessment framework that evaluates trajectories from visual observations, executed control commands, and~natural language instructions.
    \item We introduce a two-stage vision-language querying process that performs action-conditioned transition summarization followed by language-conditioned task-progress assessment.
    \item We formulate the resulting progress evaluation as an interpretable trajectory-level feedback signal that can support downstream autonomous control modules, including online off-policy policy optimization.
    \item We empirically validate the proposed system in complex 3D embodied environments and demonstrate that it provides more effective feedback than single-frame and embedding-similarity-based evaluation methods.
\end{itemize}

The remainder of this paper is organized as follows: Section~\ref{sec:related_work} reviews related work. Section~\ref{sec:preliminaries} introduces the preliminary formulations. Section~\ref{sec:methods} details the proposed \ours{} methodology. Section~\ref{sec:experiments} describes the experimental setup, and~Section~\ref{sec:results} presents the results and analysis. Finally, Sections~\ref{sec:discussion} and~\ref{sec:conclusions} provide the discussion and conclusions, respectively.

%%%%%%%%%%%%%%%%%%%%%%%%%%%%%%%%%%%%%%%%%%
\section{Related~Work} \label{sec:related_work}
\unskip

\subsection{Video Understanding for Embodied~Control}

Autonomous control systems rely on continuous sensor observations to perceive the environment, monitor state changes, and~execute sequential decisions. In~vision-based systems, these observations are naturally represented as video streams rather than isolated images. Video understanding is therefore important in embodied navigation and robotic manipulation, where the system must reason about temporal changes in the scene. Unlike static image recognition, video interpretation requires identifying object motion, interaction events, and~state transitions over time. This is particularly important for autonomous control, where visual changes are often induced by the system's own~actions.

Recent advances in vision-language models have enabled strong image-text and video-text alignment capabilities. Models such as CLIP provide general-purpose image-text representations that can be used for zero-shot recognition and task specification~\cite{radford2021learning}. More recent large vision-language models further extend multimodal reasoning and instruction-following capabilities across visual and textual inputs~\cite{zhang2024vision, alayrac2022flamingo, openai2023gpt4v, reid2024gemini, openai2024gpt4o, wang2024qwen2}. However, image-level alignment alone is often insufficient for evaluating sequential control behavior, because~task progress may depend on intermediate transitions that are not visible in a single frame. Video-language models and visual representation learning methods, including R3M, LIV, and~RoboCLIP, incorporate temporal or task-conditioned visual information and have shown promise for interpreting robot trajectories and visual demonstrations~\cite{nair2022r3m, ma2023liv, xie2018rethinking, sontakke2024roboclip}. Nevertheless, many existing methods still evaluate trajectories through embedding similarity or task-specific descriptions, which can limit their ability to explicitly relate observed visual changes to the executed action~sequence.

\subsection{Language-Guided Autonomous~Control}

Language-guided autonomous control aims to enable agents to execute tasks specified by natural language instructions. This problem has been widely studied in language-conditioned reinforcement learning and embodied instruction following, where an agent receives a language instruction and produces a sequence of actions to complete the task~\cite{tellex2011understanding, luketina2019survey, anderson2018vision, wang2019reinforced}. Benchmarks such as ALFRED and CALVIN provide visually grounded environments with natural language instructions and sequential interaction requirements, making them suitable for evaluating embodied agents in complex 3D scenes~\cite{shridhar2020alfred, kolve2017ai2, mees2022calvin}. Related work has also studied grounded navigation, manipulation, and~language-conditioned behavior learning in simulated and real environments~\cite{macmahon2006walk, kollar2010toward, misra2014dave, MacGlashan2015GroundingEC, nair2022learning}.

A major challenge in language-guided control is task evaluation. In~many environments, evaluation rules require access to privileged state information, manually engineered task logic, or~demonstrations that specify desired behavior. For~example, household manipulation tasks may require checking whether an object has been picked up, moved, cleaned, cooled, or~placed in a target receptacle. Such task-specific evaluation functions can be difficult to scale across diverse instructions and environments. Prior work has explored demonstrations, inverse reinforcement learning, and~language-conditioned reward modeling to reduce manual specification effort~\cite{ng2000algorithms, abbeel2004apprenticeship, ho2016generative, fu2019language, bahdanau2019learning}. However, these approaches often require substantial human data or task-specific supervision. In~contrast, our work focuses on generating trajectory-level task assessments directly from visual observations, executed control commands, and~natural language~instructions.

\subsection{Foundation Models for Task~Evaluation}

Foundation models have increasingly been explored as task evaluators for embodied systems. Large language models can generate task specifications or executable reward programs from natural-language descriptions~\cite{kwon2023reward, yu2023language, xie2024text2reward, ma2024eureka, wang2024robogen}. Methods such as Language to Rewards, Text2Reward, and~Eureka operate through programmatic interfaces and generally construct reward code using simulator variables, object attributes, environment APIs, or~other structured task information. Their output is an executable reward function rather than a direct assessment of a recorded visual~trajectory.

Vision-language models provide a different interface by evaluating observations directly. Existing approaches use image-text or video-text similarity as a proxy for task completion~\cite{mahmoudieh2022zero, rocamonde2024vision, ma2023liv, sontakke2024roboclip, cui2022can}, while more recent work uses large VLMs to provide success judgments or task-level feedback~\cite{du2023vision, wang2024rl}. Human-feedback research has separately studied preference-based and rating-based supervision~\cite{christiano2017deep, macglashan2017interactive, wilde2021learning, white2024rating}.

\ours{} addresses the observation-based assessment setting. It receives visual observations, the~executed action sequence, and~a natural-language instruction, and~returns a trajectory-level task-progress score without accessing privileged simulator states. Programmatic reward-generation methods are therefore discussed as related alternatives but are not included as direct baselines because they use a different evaluation interface. Our empirical comparisons are limited to the evaluated zero-shot visual and video-language baselines that can be integrated into the same observation-based control~pipeline.

\subsection{Action- and Language-Conditioned Video~Assessment}

Trajectory evaluation for embodied control should account for both the executed action sequence and the task instruction. A~video trajectory is not merely a collection of frames: the actions provide temporal context for interpreting the observed transitions, while the instruction defines which transitions constitute task-relevant progress. We therefore describe the target problem as action- and language-conditioned video~assessment.

Prior video reasoning research shows that models can struggle to interpret temporal relationships and action-dependent events beyond static scene recognition~\cite{xiao2021nextqa, li2022causalvidqa, zang2023discovering, chen2024mecd}. These studies are primarily formulated as video question answering or event-structure analysis, whereas the present work focuses on assigning task-progress feedback to embodied-control~trajectories.

\ours{} addresses this trajectory-assessment problem through two stages: action-conditioned transition summarization and language-conditioned task-progress assessment. Static image-text matching can indicate whether a final frame resembles a target description, while embedding-based video similarity can capture temporal patterns. In~contrast, the~proposed decomposition produces an explicit summary of the action-to-transition correspondence before assigning a discrete task-progress~score.

%%%%%%%%%%%%%%%%%%%%%%%%%%%%%%%%%%%%%%%%%%
\section{Preliminaries} \label{sec:preliminaries}
\unskip

\subsection{Action-Conditioned Video~Trajectories}

Let $I_t \in \mathcal{I}$ denote the RGB observation at time step $t$, $u_t \in \mathcal{U}$ the executed control command, and~$l \in \mathcal{L}$ the natural-language instruction. We represent an interaction of length $T$ as an action-conditioned video trajectory
\begin{equation}
\tau=(I_0,u_0,I_1,u_1,\ldots,I_{T-1},u_{T-1},I_T).
\label{eq:trajectory}
\end{equation}
The trajectory preserves the temporal order of the visual observations and the corresponding executed~commands.

\subsection{Action- and Language-Conditioned Task~Assessment}

We define the trajectory evaluator as
\begin{equation}
F:\mathcal{T}\times\mathcal{L}\rightarrow\mathcal{R},
\label{eq:evaluator}
\end{equation}
where $\mathcal{R}=\{0,1,\ldots,k-1\}$ is a discrete task-progress scale, and~$\mathcal{T}$ is the space of such trajectories. In~the experiments, $k=4$: a score of 0 denotes no task-relevant progress and a score of 3 denotes full task completion. The~evaluator uses the visual transitions, executed commands, and~instruction rather than relying only on the final~observation.

\subsection{Assessment-Driven Policy~Optimization}

The evaluator output is used as terminal feedback for downstream policy optimization. For~a collected trajectory $\tau$ and instruction $l$, intermediate rewards are set to zero and the final reward is assigned as
\begin{equation}
r_{T-1}=F(\tau,l).
\label{eq:terminal_feedback}
\end{equation}
This formulation connects trajectory-level task-progress assessment to downstream control without requiring a manually engineered task-specific reward function. The~policy-optimization algorithm itself is not a contribution of this~work.

\subsection{Vision-Language Models as Trajectory~Evaluators}

We instantiate $F$ with a pre-trained vision-language model. The~visual prompt constructor $\Omega$ maps sampled observations from $\tau$ into a composite visual input. The~summarization prompt constructor $\Psi_S$ asks the model to describe action-conditioned visual transitions, and~the task-progress assessment prompt constructor $\Psi_C$ asks it to assign a discrete score from the generated summary. Section~\ref{sec:two_stage} specifies this two-stage~process.

\section{Materials and~Methods} \label{sec:methods}

\ours{} instantiates the autonomous evaluator $F(\tau,l)$ defined in the Preliminaries through a structured two-stage VLM querying pipeline. Given an action-conditioned video trajectory $\tau$ and a language instruction $l$, the~objective is to infer a trajectory-level feedback score $r \in \mathcal{R}$ directly from visual observations and executed control commands. The~framework uses a pre-trained VLM $H$ as the core trajectory evaluator, together with a visual prompt constructor $\Omega$, a~summarization prompt constructor $\Psi_S$, and~a task-progress assessment prompt constructor $\Psi_C$. This design allows the system to approximate $F(\tau,l)$ without manually engineered task-specific evaluation~rules.

\begin{figure}[H]
    
    \includegraphics[width=\textwidth]{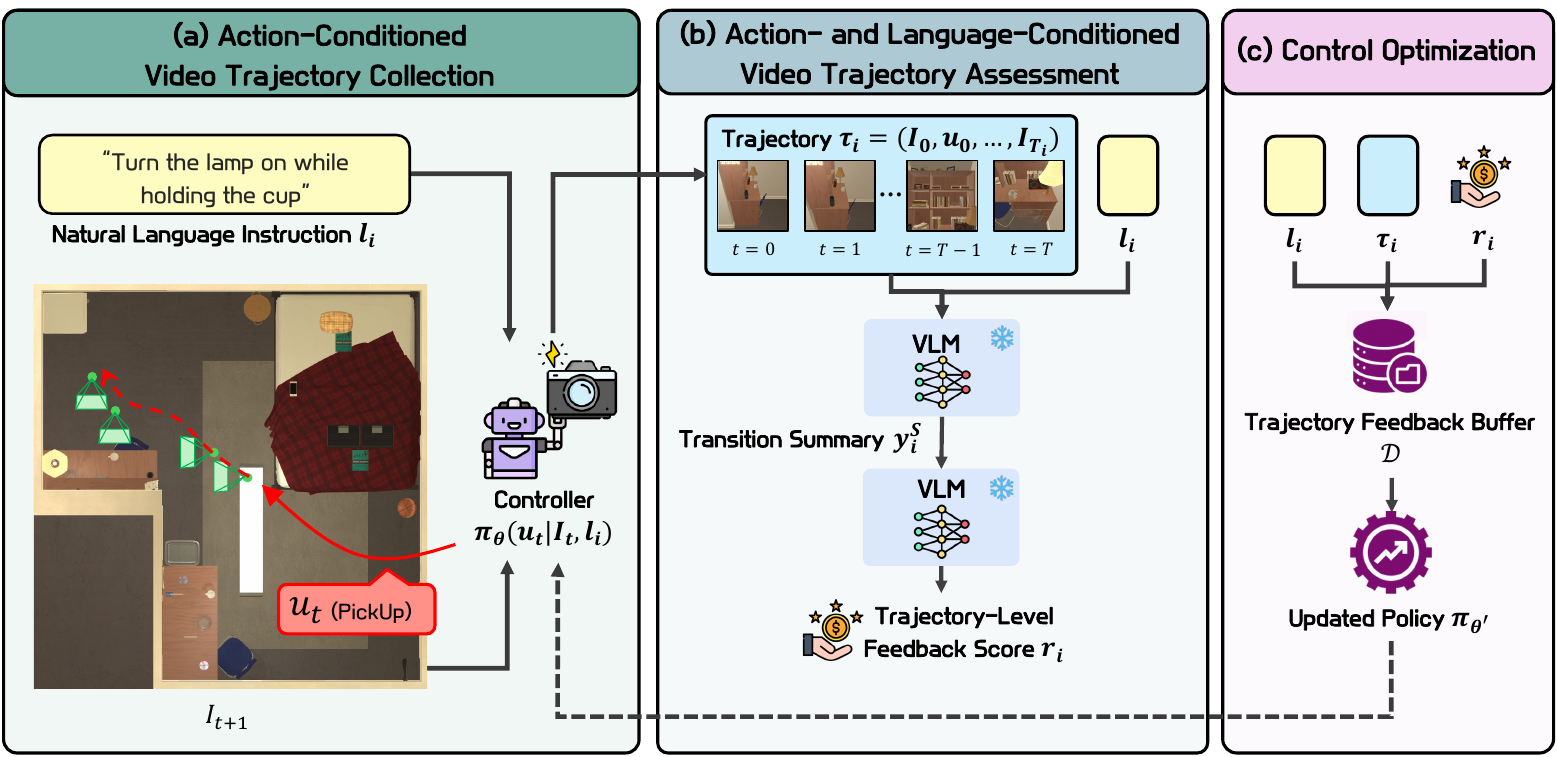}
    \caption{Overview of \ours{}.
    (\textbf{a}) A language-conditioned controller collects an action-conditioned video by executing control commands from visual observations under a natural language instruction.
    (\textbf{b}) A two-stage VLM trajectory evaluator first summarizes action-induced visual transitions from the collected stream and then estimates task progress with respect to the instruction. The snowflake symbol indicates that the pre-trained VLM is frozen and its parameters are not updated during control optimization.
    (\textbf{c}) The resulting trajectory-level feedback score is stored in the evaluated trajectory buffer and used for assessment-driven control optimization.
    }
    \label{fig:overview}
\end{figure}

The overall procedure is illustrated in Figure~\ref{fig:overview} and summarized in Algorithm~\ref{alg:pseudo_code}. The~controller first interacts with the environment to collect an action-conditioned video trajectory. The~collected trajectory is then converted into a composite visual input and a structured text prompt. In~the first stage, the~VLM summarizes action-induced visual transitions. In~the second stage, the~generated summary is evaluated with respect to the instruction to produce a discrete task-progress score. The~resulting score is directly used as a trajectory-level feedback signal for downstream autonomous~control.

\begin{algorithm}[H]
    \small
    \caption{\ours{} closed-loop evaluation and optimization~procedure.}
    \label{alg:pseudo_code}
    \begin{algorithmic}[1]
        \STATE \textbf{Input}: pre-trained VLM $H$, visual prompt constructor $\Omega$, prompt constructors $\Psi_S$ and $\Psi_C$
        \STATE \textbf{Initialize}: Language-conditioned controller $\pi_{\theta}$, trajectory buffer $\mathcal{D}$
        \WHILE{not converged}
            \STATE Sample a language instruction $l_i \sim \mathcal{L}$
            \STATE Run $\pi_{\theta}$ to collect trajectory $\tau_i=(I_0,u_0,\ldots,I_{T_i-1},u_{T_i-1},I_{T_i})$
            \STATE Construct summarization inputs: $I_i^{\mathrm{vis}}=\Omega(\{I_t\}_{t=0}^{T_i})$, $p_i^S=\Psi_S(\{u_t\}_{t=0}^{T_i-1},l_i)$
            \STATE Query for action-conditioned transition summary: $y_i^S=H(p_i^S,I_i^{\mathrm{vis}})$
            \STATE Construct assessment prompt: $p_i^C=\Psi_C(y_i^S,l_i)$
            \STATE Query for trajectory-level feedback: $r_i=H(p_i^C, \emptyset)$
            \STATE Stored evaluated trajectory: $\mathcal{D}\leftarrow\mathcal{D}\cup\{(\tau_i,l_i,y_i^S,r_i)\}$
            \STATE Update $\pi_{\theta}$ using data sampled from $\mathcal{D}$ with an off-policy optimization algorithm
        \ENDWHILE
    \end{algorithmic}
\end{algorithm}

This procedure differs from approaches that first learn a separate feedback prediction model or evaluation model from human feedback or demonstrations~\cite{macglashan2017interactive, wilde2021learning, white2024rating}. Such intermediate modeling steps can introduce additional complexity and may be vulnerable to feedback misspecification or misgeneralization~\cite{casper2023open}. In~contrast, \ours{} queries the VLM directly as a trajectory-level evaluator and uses the resulting score as feedback. The~central contribution is therefore not a new policy optimization algorithm, but~an action- and language-conditioned video assessment mechanism that produces interpretable trajectory-level signals from visual observations and control~histories.

\subsection{Two-Stage Action- and Language-Conditioned Video~Assessment} \label{sec:two_stage}

Natural language instructions in autonomous control scenarios can be highly compositional. For~example, an~instruction such as ``Put the coffee cup in the sink, turn on the water, turn off the water, and~pick up the coffee cup'' involves multiple sequential sub-tasks. A~reliable evaluator must therefore assess not only the final visual observation, but~also the intermediate action-induced visual transitions that occur throughout the trajectory. This is particularly important because multiple valid control strategies may lead to the same task completion state, while failed trajectories can still resemble successful ones in the final~frame.

\begin{figure}[H]
    
    \includegraphics[width=\textwidth]{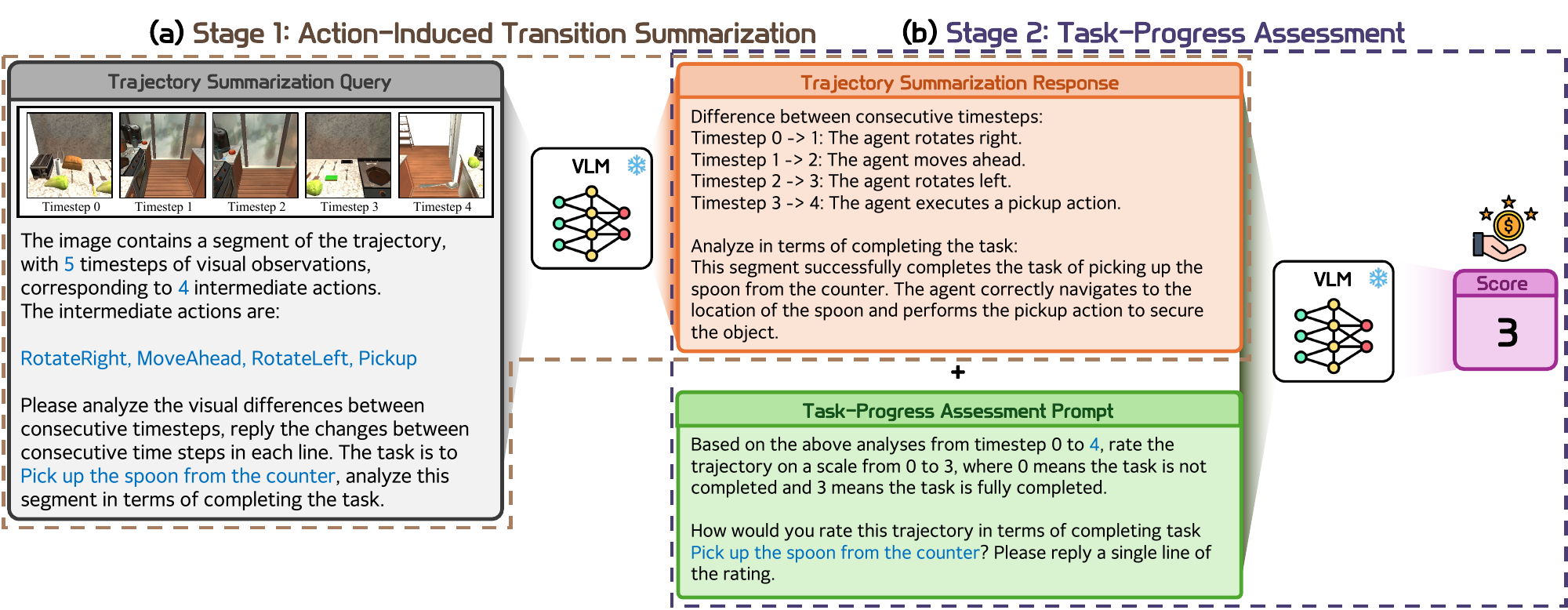}
    \caption{Illustration of the two-stage VLM querying process used in \ours{}. Blue text indicates task- and trajectory-specific variables that are replaced according to the input. The snowflake symbol indicates that the pre-trained VLM is frozen and its parameters are not updated during control optimization.
    (\textbf{a}) In Stage 1, Action-Induced Transition Summarization, sampled visual observations are arranged into a composite visual prompt with timestep annotations, and~the executed control commands are provided as textual context. The~VLM then summarizes frame-to-frame visual transitions conditioned on the action sequence and the language instruction.
    (\textbf{b}) In Stage 2, Task-Progress Assessment, the~generated transition summary is combined with a rating prompt and evaluated with respect to the original instruction. The~VLM outputs a discrete trajectory-level feedback score, where higher scores indicate greater task progress toward full completion.
    }
    \label{fig:prompt}
\end{figure}

To address this issue, \ours{} evaluates the full action-conditioned video trajectory rather than a single observation. As~illustrated in Figure~\ref{fig:prompt}, the~two-stage querying process separates transition interpretation from task-progress assessment. The~first stage asks the VLM to summarize how the visual scene changes between consecutive observations, conditioned on the executed control commands and the language instruction. The~second stage then uses this generated summary to infer whether the observed visual transitions constitute meaningful progress toward completing the instruction. This decomposition makes the assessment more interpretable and reduces the burden of asking the VLM to directly infer task completion from raw visual inputs~alone.

Concretely, given an action-conditioned video trajectory $\tau_i$, the~visual prompt constructor $\Omega$ maps the sampled visual observations into a composite visual input:
\begin{equation}
I_i^{\mathrm{vis}}=\Omega(\{I_t\}_{t=0}^{T_i}).
\label{eq:visual_prompt}
\end{equation}
The composite input is constructed by concatenating sampled video frames and adding timestep annotations to preserve the temporal order. Such visual prompting strategies are related to prior studies showing that structured visual prompts can improve the reasoning behavior of pre-trained vision-language models~\cite{jia2022visual, bar2022visual, shtedritski2023does}.

The summarization prompt constructor $\Psi_S$ maps the executed control commands and the instruction into a text prompt:
\begin{equation}
p_i^S=\Psi_S(\{u_t\}_{t=0}^{T_i-1},l_i).
\label{eq:summary_prompt}
\end{equation}
This prompt includes the sequence length, the~executed command list, and~an instruction asking the VLM to describe frame-to-frame visual changes in relation to the task. The~VLM then generates an action-conditioned trajectory summary:
\begin{equation}
y_i^S=H(p_i^S,I_i^{\mathrm{vis}}).
\label{eq:summary_generation}
\end{equation}
The summary $y_i^S$ serves as an interpretable intermediate representation that describes the action-to-visual transitions observed in the~trajectory.

For the second stage, the~task-progress assessment prompt constructor $\Psi_C$ takes the generated summary and the original instruction as input:
\begin{equation}
p_i^C=\Psi_C(y_i^S,l_i).
\label{eq:assessment_prompt}
\end{equation}
The VLM is then queried to produce a discrete trajectory-level feedback score:
\begin{equation}
r_i=H(p_i^C, \emptyset), \qquad r_i\in\mathcal{R}=\{0,1,\ldots,k-1\}.
\label{eq:vlm_feedback_score}
\end{equation}
In our experiments, we use $k=4$, corresponding to the progress scale $\{0,1,2,3\}$, where 0 denotes no task-relevant progress and 3 denotes full task completion. This score is used directly as the feedback signal for downstream autonomous control. Unlike comparative feedback, which only indicates relative preference between trajectory pairs, this evaluative score provides an absolute assessment of task progress, which is better suited for multi-step instruction-following scenarios~\cite{wilde2021learning, white2024rating}.

\subsection{Implementation~Details}

Action-conditioned video trajectories can contain up to 50 interaction steps in the evaluated environments. To~avoid constructing an excessively large composite input, we divide long trajectories into temporal segments and summarize each segment separately. Unless~otherwise specified, the~segment length is 10 frames. The~segment-level summaries are concatenated and passed to the second-stage task-progress~assessment.

The assessment-stage input is textual because it contains the generated transition summary and the original instruction. We use the same VLM backbone for both stages for implementation consistency. The~experiments use the model aliases reported in Section~\ref{sec:experiments}; each alias was held fixed within an evaluation run. Provider-default decoding settings were~used.

A task-progress response is accepted only when the complete stripped response is exactly one integer in $\{0,1,2,3\}$. An~empty response, a~response without a valid integer, or~a value outside this range is treated as an invalid evaluator output. When this occurs during closed-loop optimization, no evaluator-derived feedback is stored for that trajectory and the corresponding evaluator-driven update is skipped. The~current study does not evaluate sensitivity to prompt paraphrases or repeated~queries.

Detailed downstream policy-optimization settings, environment configurations, and~task lists are provided in Appendices~\ref{app:implementation_environment} and~\ref{app:task_details}.

\section{Experiments} \label{sec:experiments}

We conduct experiments to evaluate the effectiveness and reliability of the proposed \ours{} framework in visually grounded 3D embodied environments. The~experiments are conducted in four household workspaces: Kitchen, Bathroom, Living Room, and~Bedroom, using environments based on the ALFRED benchmark~\cite{shridhar2020alfred} and the AI2-THOR simulator~\cite{kolve2017ai2}. Each environment requires an autonomous agent to complete natural language instructions through sequential interaction with visual observations and executed control~commands.

Our experimental design has two main objectives. First, we evaluate whether \ours{} can reliably assess action-conditioned video trajectories by comparing its trajectory-level feedback scores with ground-truth task completion labels. Second, we examine whether the generated feedback can support downstream autonomous control when used in a closed-loop setting. This design allows us to analyze both the diagnostic reliability of the trajectory evaluator and its utility as a feedback mechanism for language-guided autonomous~control.

\subsection{Experimental~Setup}

\begin{figure}[H]
    
    \includegraphics[width=\textwidth]{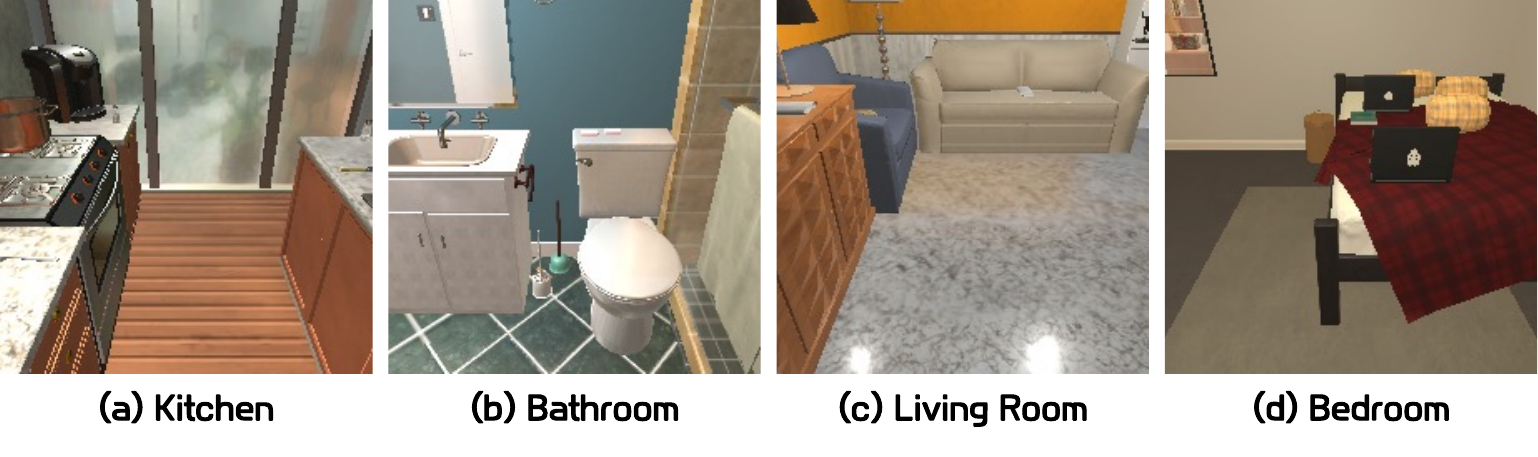}
    \caption{Representative egocentric views of the four unseen AI2-THOR floor plans used in the ALFRED-based~experiments.}
    \label{fig:alfred}
\end{figure}

We evaluate \ours{} on a set of visually grounded household tasks sampled from four ALFRED environments: Kitchen, Bathroom, Living Room, and~Bedroom~\cite{shridhar2020alfred}. Each task consists of two sequential sub-tasks specified by natural language instructions. The~agent receives RGB visual observations from an embodied camera and executes discrete navigation and interaction commands. An~episode is considered successful only when both sub-tasks are completed. Representative egocentric views of the four evaluation environments are shown in Figure~\ref{fig:alfred}. Further details on the environment configuration, action space, and~task instructions are provided in Appendices~\ref{app:implementation_environment} and~\ref{app:task_details}.

For the VLM-based evaluator, we use the Gemini 1.5 Pro model alias as the primary vision-language model~\cite{reid2024gemini}. Given an action-conditioned video trajectory and a language instruction, \ours{} generates a discrete trajectory-level feedback score $r \in \{0,1,2,3\}$ through the two-stage querying process described in Section~\ref{sec:two_stage}. A~score of 0 indicates no task-relevant progress, while a score of 3 indicates full task completion. Unless~otherwise specified, we use a temporal segment length of 10 frames during the first-stage video summarization~process.

\subsection{Baseline~Evaluators}
To evaluate the proposed action- and language-conditioned assessment framework, we compare \ours{} against three zero-shot visual evaluation baselines—CLIP, R3M, and~RoboCLIP—and include a ground-truth oracle as an upper-bound~reference.

\begin{itemize}
    \item CLIP~\cite{radford2021learning}:
    A widely used vision-language foundational model. In~our experiments, the~CLIP-based evaluator computes the cosine similarity between the final visual observation $I_T$ and the natural-language instruction $l$. This serves as a standard static, single-frame image-text matching baseline that does not explicitly model temporal~transitions.
    
    \item R3M~\cite{nair2022r3m}: A visual representation model pre-trained on the Ego4D human video dataset to capture robust temporal dynamics and human--object interactions. The~R3M evaluator generates feedback by measuring the embedding similarity between the visual observations and the task context. While it incorporates visual information relevant to robotic control, it lacks the ability to explicitly summarize action-induced visual~transitions.
    
    \item RoboCLIP~\cite{sontakke2024roboclip}: A video-language representation model for embodied control. It generates a trajectory-level feedback score by computing the multimodal alignment score between the continuous video sequence $\tau$ and the language instruction $l$. Although~it incorporates temporal visual information, the~feedback is still derived from continuous embedding similarity rather than an explicit interpretation of action-to-visual~transitions.

    \item Ground-Truth Oracle (Oracle): This upper-bound reference uses the environment-provided ground-truth task completion signal. It is not available in uninstrumented, open-ended autonomous systems and is used solely to contextualize the performance gap between the evaluation-driven policies and the environment-provided upper-bound reference.
\end{itemize}

To ensure a fair and isolated comparison, all baseline evaluators are integrated into the identical downstream closed-loop control pipeline with Implicit Q-learning (IQL)~\cite{kostrikov2022offline} without any human-engineered reward functions. The~controller utilizes the similarity scores generated by these baselines directly as the optimization feedback, which is then evaluated against the discrete, action- and language-conditioned assessment signal $r$ provided by~\ours{}.

\subsection{Trajectory Data~Collection}

% To analyze the assessment reliability of \ours{} independently from the closed-loop optimization process, we first collect a dataset of action-conditioned video trajectories. Following prior work on ALFRED-based autonomous control~\cite{zhang2023bootstrap}, we deploy controllers at different optimization stages to generate trajectories with varying degrees of task completion.
% % For each environment, we collect approximately 500 trajectories. Since each task consists of two sequential sub-tasks, we summarize collection quality using the average ground-truth return, defined as the fraction of completed sub-tasks per trajectory, which lies between 0.6 and 0.7. This return is distinct from the binarized full-task success label used for the confusion matrices, where a trajectory is counted as positive only when both sub-tasks are completed; the proportion of fully successful trajectories is therefore lower than the average return. The resulting dataset spans successful, partially successful, and failed trajectories.
% For each environment, we collect approximately 500 trajectories by deploying controllers at different optimization stages, which yields a mixture of successful, partially successful, and failed trajectories. For the confusion-matrix analysis, each trajectory is binarized using its ground-truth label: a trajectory is counted as positive only when both sub-tasks are completed, and negative otherwise.

To analyze the assessment reliability of \ours{} independently from the closed-loop optimization process, we first collect a dataset of action-conditioned video trajectories. Following prior work on ALFRED-based autonomous control~\cite{zhang2023bootstrap}, we deploy controllers at different optimization stages to generate approximately 500 trajectories per environment, yielding a mixture of successful, partially successful, and~failed trajectories. For~the confusion-matrix analysis, each trajectory is binarized using its ground-truth label: it is counted as positive only when both sub-tasks are completed, and~negative~otherwise.

Each trajectory is processed by \ours{} to obtain a discrete feedback score $r \in \{0,1,2,3\}$. To~compare the evaluator output with the binary ground-truth task completion label, we binarize the score as follows: $r=3$ is treated as a positive prediction, indicating full task completion, while $r<3$ is treated as a negative prediction. This binarization is used only for the diagnostic confusion-matrix analysis. Closed-loop optimization uses the original score in $\{0,1,2,3\}$ directly as terminal feedback, so intermediate progress scores are not~discarded.

\subsection{Evaluation~Metrics}

We use standard classification metrics to evaluate the diagnostic reliability of trajectory-level task assessment. Let true positives denote successful trajectories correctly assigned the maximum feedback score, and~false positives denote failed or incomplete trajectories incorrectly assigned the maximum~score.

\begin{itemize}
    \item Accuracy (Acc.): The proportion of correctly classified successful and unsuccessful trajectories among all evaluated~trajectories.

    \item Precision (Prec.): The proportion of predicted successful trajectories that are actually successful. High precision indicates that the evaluator rarely assigns a false success signal to an incomplete~trajectory.

    \item Recall (Rec.): The proportion of ground-truth successful trajectories correctly identified by the evaluator. High recall indicates that the evaluator detects most successful~executions.

    \item F1-Score (F1): The harmonic mean of precision and recall, providing a balanced measure of evaluation reliability.
\end{itemize}

In the context of autonomous control, precision is particularly important because false positive feedback can reinforce incorrect behavior. However, recall is also relevant because an overly conservative evaluator may fail to reward successful trajectories. Therefore, we report all four metrics to characterize both the precision-recall trade-off and sensitivity of the proposed evaluation~system.

\subsection{Closed-Loop Feedback~Evaluation}

In addition to offline reliability analysis, we evaluate whether the trajectory-level feedback generated by \ours{} can support downstream autonomous control. The~generated feedback score is converted into the terminal reward signal for online off-policy optimization. We use IQL~\cite{kostrikov2022offline} as the downstream optimization algorithm, following prior work on ALFRED-like embodied environments~\cite{zhang2023bootstrap}. The~purpose of this experiment is not to propose a new policy optimization algorithm, but~to test whether action- and language-conditioned video assessment produces feedback that is useful for improving autonomous task~execution.

For each method, we train the controller using the corresponding feedback source and evaluate task completion rates over training. Performance is measured every fixed number of optimization iterations and averaged over multiple evaluation episodes. We report both learning curves and maximum task completion rates across the four~environments.

%%%%%%%%%%%%%%%%%%%%%%%%%%%%%%%%%%%%%%%%%%
\section{Results} \label{sec:results}

We evaluate the \ours{} framework in three parts. Sections~\ref{sec:conservative_assessment}--\ref{sec:latency} analyze diagnostic reliability, temporal-window trade-offs, and~VLM latency. Section~\ref{sec:policy_optimization} evaluates the use of the resulting scores as closed-loop terminal feedback. Section~\ref{sec:ablation} examines the contributions of temporal annotations, executed actions, and~evaluative~scoring.

\subsection{Analysis of Conservative Action-Conditioned~Assessment} \label{sec:conservative_assessment}
\begin{figure}[H]

\includegraphics[width=\textwidth]{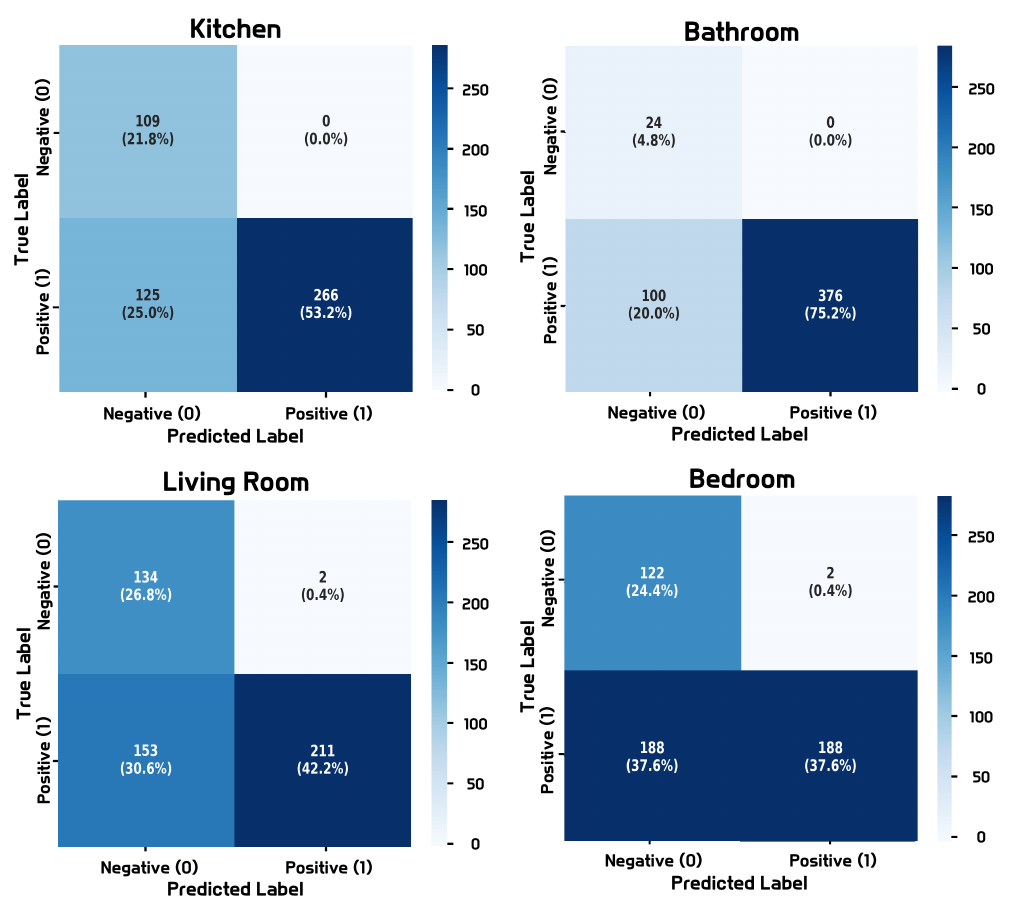}
\caption{Trajectory assessment confusion matrices across four task domains using the primary VLM evaluator. The~matrices highlight the near-zero false-positive behavior of the proposed evaluator, with~false-positive rates below 1\% in all evaluated~environments. \label{fig:sensor_confusion_matrices}}
\end{figure}

We first analyze the operating behavior and diagnostic characteristics of our primary VLM evaluator (Gemini 1.5 Pro). To~quantify the evaluator's reliability, we map its binarized predictions across tasks using domain-specific $2 \times 2$ confusion matrices, as~illustrated in Figure~\ref{fig:sensor_confusion_matrices}.

\vspace{-6pt}

%\unskip

The matrices reveal a highly conservative action-conditioned assessment characteristic across all workspaces. The~system consistently achieves a precision approaching 100\%, demonstrating that it rarely assigns the maximum score to failed trajectories. Reducing false-positive feedback is important because erroneous success signals may reinforce unsuccessful behavior. Conversely, recall is lower than accuracy and precision because the VLM frequently assigns a score of 2 rather than 3 to successful trajectories when the correspondence between the executed action and the observed visual transition is ambiguous. This conservative behavior reduces false-positive feedback but fails to identify some successful~trajectories.

\subsection{Temporal Window~Optimization} \label{sec:temporal_window}
\begin{figure}[H]
%\centering
\includegraphics[width=\textwidth]{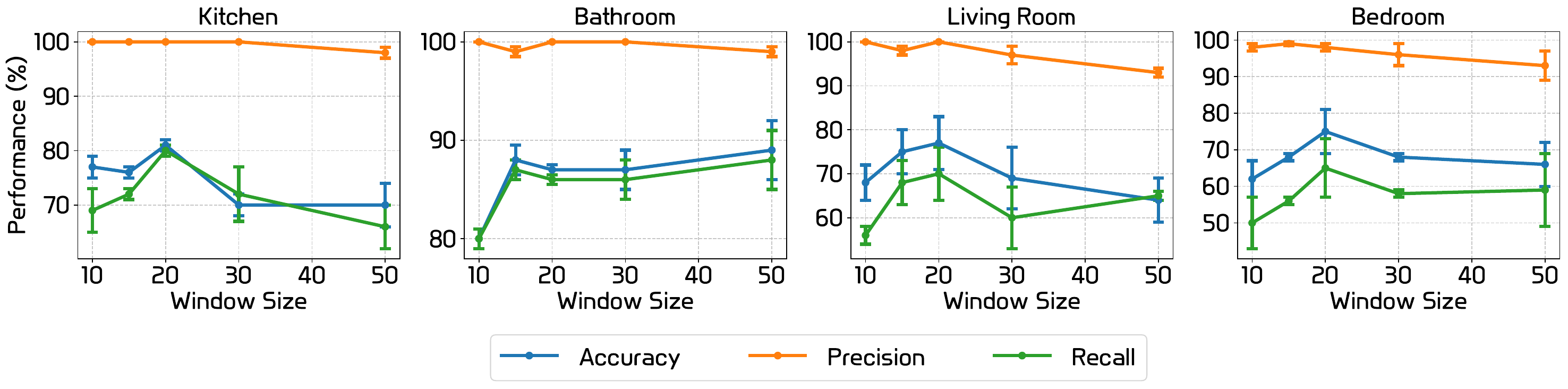}
\caption{Impact of the temporal window size (segment length) on action-conditioned assessment reliability. The~analysis demonstrates a system trade-off between contextual fragmentation (narrow windows) and visual input downsampling and increased inference cost (wide windows).\label{fig:temporal_window_optimization}}
\end{figure}

We examine how the temporal segment length affects trajectory-assessment reliability. Figure~\ref{fig:temporal_window_optimization} tracks the performance metrics (accuracy, precision, and~recall) across different temporal window~sizes.

%\unskip

The results indicate a clear system trade-off regarding the temporal resolution. If~the segment length is too narrow, the~temporal context is broken, hindering the model's ability to reason about action-conditioned visual transitions. Conversely, extending the window size beyond 20 frames causes a degradation in performance due to visual input downsampling and increased reasoning difficulty for large composite prompts. A~segment length of 20 achieves the strongest diagnostic reliability in the offline assessment analysis. However, we use a segment length of 10 in the closed-loop control experiments to reduce inference cost while maintaining competitive assessment~performance.

\subsection{VLM Backbone Benchmarking and Latency~Trade-Offs} \label{sec:latency}
For practical closed-loop control applications, system latency is as critical as accuracy. We benchmark various VLM backbones, categorizing them into larger VLMs (Gemini 1.5 Pro, GPT-4o, Qwen2-VL-72B) and lightweight VLMs (Gemini 1.5 Flash, GPT-4o mini).

\begin{figure}[H]

\includegraphics[width=\textwidth]{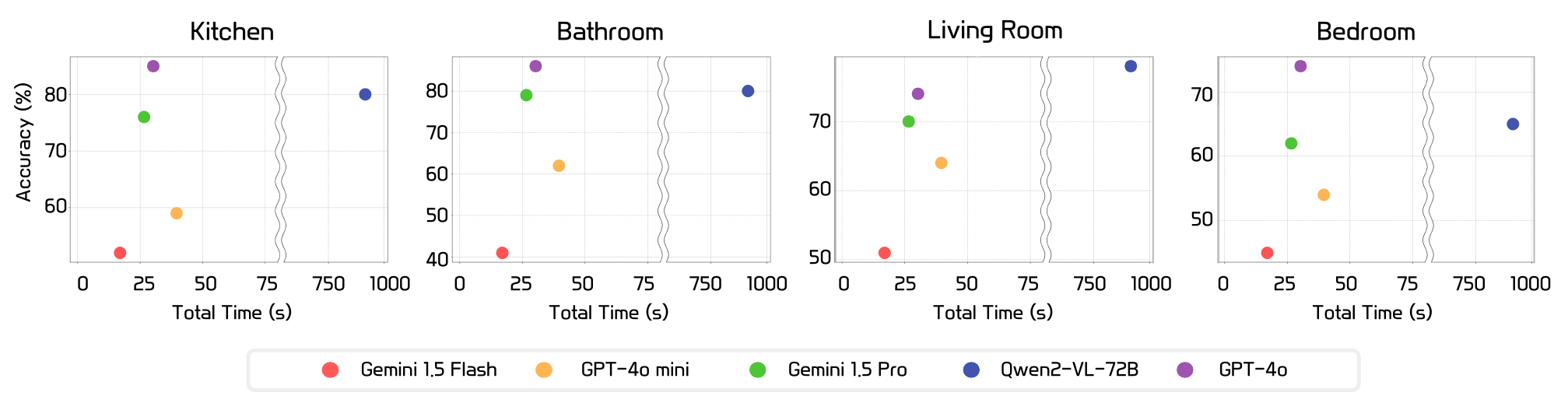}
\caption{System feasibility trade-off analysis mapping end-to-end computational latency against overall evaluation reliability. Larger VLMs generally occupy the higher-accuracy, higher-latency region, whereas lightweight VLMs provide lower-latency but less reliable~assessment. \label{fig:pareto_latency_reliability}}
\end{figure}

Figure~\ref{fig:pareto_latency_reliability} illustrates the trade-offs between total execution latency and evaluation accuracy. Lightweight VLMs have lower latency but exhibit lower accuracy and precision. Flagship-tier models provide higher reliability but require longer inference times. Crucially, our system latency profiling reveals that over 90\% of the end-to-end processing delay occurs during the Stage 1 (Action-Induced Transition Summarization) phase (see Table~\ref{tab:stage_latency} in Appendix~\ref{app:implementation_environment} for the per-stage breakdown), whereas the Stage 2 (Task-Progress Assessment), which processes compressed semantic text, contributes negligibly to the overall overhead. These measurements show that Stage 1 is the principal computational bottleneck. Gemini 1.5 Pro provides a favorable accuracy-latency trade-off among the evaluated backbones, but~its approximately 25-second Stage 1 latency remains unsuitable for high-frequency real-time~control.
\vspace{-6pt}

\subsection{Assessment-Driven Policy~Optimization} \label{sec:policy_optimization}
\begin{figure}[H]

\includegraphics[width=\textwidth]{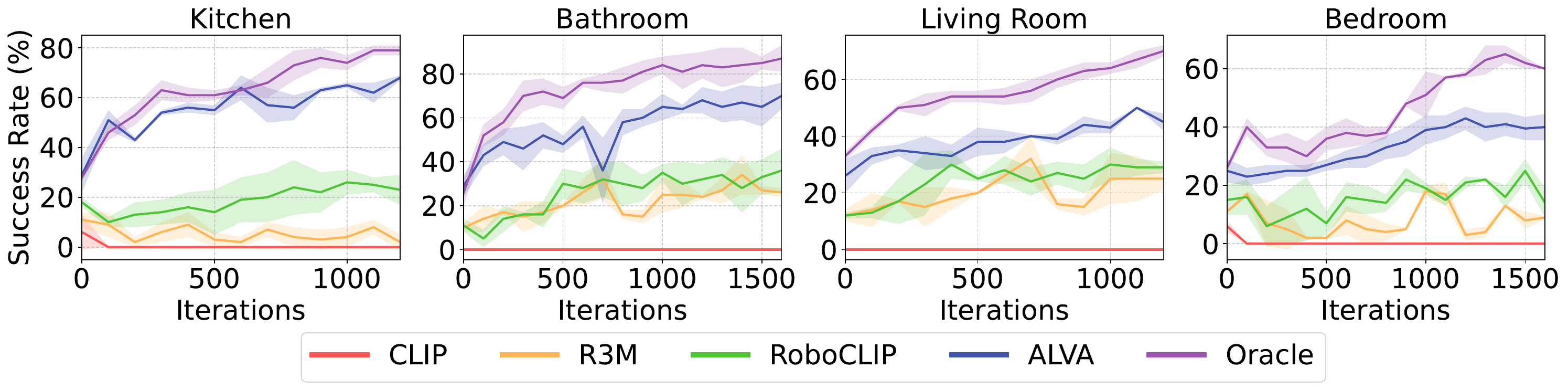}
\caption{Closed-loop autonomous control performance across four simulated 3D embodied workspaces. The~curves track the task completion rate over continuous control optimization iterations for Kitchen, Bathroom, Living Room, and~Bedroom environments. The~solid lines denote the mean performance, while the shaded regions indicate the standard~deviation.\label{fig:control_learning_curves}}
\end{figure}

We next evaluate whether the trajectory-level scores can support downstream policy optimization. Figure~\ref{fig:control_learning_curves} shows task completion rates over optimization~iterations. 

%\unskip

To summarize the best observed policy performance during optimization, Table~\ref{tab:maximum_success_rates} reports the maximum task-completion rate achieved by each~method.

\begin{table}[H] 
\caption{Maximum task completion rates achieved across different evaluation-driven control models within the four simulated~workspaces. Bold indicates the best performance among the non-oracle methods. \label{tab:maximum_success_rates}}

\begin{tabularx}{\textwidth}{l C C C C C}
\toprule
\textbf{Model} & \textbf{Kitchen} & \textbf{Bathroom} & \textbf{Living Room} & \textbf{Bedroom} & \textbf{Average} \\
\midrule
Ground-Truth Oracle  & 79.0\stdv{2.0} & 87.0\stdv{6.0} & 70.0\stdv{2.0} & 65.0\stdv{3.0} & 75.3\stdv{3.3} \\
\midrule
CLIP~\cite{radford2021learning}         & 6.0\stdv{7.0}  & 0.0\stdv{0.0}  & 0.0\stdv{0.0}  & 6.0\stdv{2.0}  & 3.0\stdv{2.3} \\
R3M~\cite{nair2022r3m}                  & 11.0\stdv{3.0} & 34.0\stdv{9.0} & 32.0\stdv{8.0} & 18.0\stdv{2.0} & 23.8\stdv{5.5} \\
RoboCLIP~\cite{sontakke2024roboclip}    & 26.0\stdv{5.0} & 36.0\stdv{10.0}& 30.0\stdv{5.0} & 25.0\stdv{4.0} & 29.3\stdv{6.0} \\
\midrule
\ours{} (Ours)                 & \textbf{68.0\stdv{1.0}} & \textbf{70.0\stdv{6.0}} & \textbf{50.0\stdv{0.0}} & \textbf{43.0\stdv{4.0}} & \textbf{57.8\stdv{2.8}} \\
\bottomrule
\end{tabularx}
\end{table}
%\unskip

In the evaluated simulated tasks, \ours{} achieves higher maximum task-completion rates than CLIP, R3M, and~RoboCLIP and reduces the gap to the ground-truth oracle. These results are limited to the evaluated zero-shot visual and video-language baselines under the shared observation-based control~pipeline.

\subsection{Ablation Study on Component Configurations and Feedback~Types} \label{sec:ablation}
To understand the architectural drivers behind the success of the \ours{} framework demonstrated above, we conduct an ablation study. We decompose this analysis into two parts: evaluating the impact of data fusion configurations (Table~\ref{tab:ablation_data_fusion}), and~comparing structural feedback paradigms (Table~\ref{tab:ablation_feedback}).

\begin{table}[H] 
\caption{Ablation analysis tracking evaluation reliability across different data fusion configurations. All configurations utilize Evaluative scoring to isolate the impact of `Temporal' (timestep annotations) and `Control' (executed action logs) inputs. F1 is averaged across environments and therefore need not equal the harmonic mean of the reported mean precision and recall. Bold indicates the best performance among methods.
\label{tab:ablation_data_fusion}}
\centering
\begin{tabularx}{\textwidth}{l  C C C C}
\toprule
\textbf{Configuration} & \textbf{Acc. (\%)} & \textbf{Prec. (\%)} & \textbf{Rec. (\%)} & \textbf{F1 (\%)} \\
\midrule
Video Only & 45.3\stdv{3.3} & 92.3\stdv{2.3} & 35.5\stdv{2.8} & 50.8\stdv{3.9} \\
\quad + Temporal & 48.3\stdv{3.5} & 97.3\stdv{1.3} & 38.3\stdv{3.8} & 54.5\stdv{4.4} \\
\quad + Control & 64.8\stdv{4.0} & 99.3\stdv{0.5} & 56.0\stdv{3.8} & 70.9\stdv{3.4} \\
\quad + Temporal + Control (Ours) & \textbf{71.5\stdv{3.5}}
 & \textbf{99.5\stdv{0.8}} & \textbf{63.8\stdv{3.8}} & \textbf{77.2\stdv{3.4}} \\
\bottomrule
\end{tabularx}
\end{table}
%\unskip

The video-only configuration provides lower evaluation reliability than the configurations that include executed actions. Adding the action sequence produces the largest improvement among the evaluated structural inputs, indicating that action context helps the VLM relate visual changes to the agent's executed~commands. 

\begin{table}[H] 
\caption{Performance comparison between preference-based and evaluative feedback paradigms. Both models utilize the fully integrated data~stream. Bold indicates the best performance among methods. \label{tab:ablation_feedback}}
\centering
\begin{tabularx}{\textwidth}{l C C C C}
\toprule
\textbf{Feedback Type} & \textbf{Acc. (\%)} & \textbf{Prec. (\%)} & \textbf{Rec. (\%)} & \textbf{F1 (\%)} \\
\midrule
Preference-based       & 38.3\stdv{5.5}         & 88.8\stdv{4.3}          & 27.5\stdv{6.3}       & 41.8\stdv{4.7} \\
Evaluative (Ours) & \textbf{71.5\stdv{3.5}} & \textbf{99.5\stdv{0.8}} & \textbf{63.8\stdv{3.8}} & \textbf{77.2\stdv{3.4}} \\
\bottomrule
\end{tabularx}
\end{table}
%\unskip

Table~\ref{tab:ablation_feedback} shows that evaluative scoring yields higher diagnostic metrics than preference-based feedback under the fully integrated input configuration. In~this experiment, the~discrete rating provides an absolute estimate of task progress, whereas pairwise preference feedback does not directly encode the degree of~completion.

%%%%%%%%%%%%%%%%%%%%%%%%%%%%%%%%%%%%%%%%%%
\section{Discussion} \label{sec:discussion}

The experiments show that action- and language-conditioned video assessment can provide useful trajectory-level feedback for the evaluated simulated household tasks. Compared with final-frame image-text matching, \ours{} first interprets visual transitions in the context of the executed action sequence and then estimates progress toward the instruction. The~ablation results indicate that both action context and temporal annotations improve assessment reliability. However, these findings are limited to the evaluated environments, tasks, prompts, and~VLM~backbones.

\subsection{Limitations}

The conservative scoring behavior of \ours{} reduces false-positive feedback at the cost of lower recall, as~some successful trajectories receive scores below the maximum. Although~closed-loop optimization does not use the binary score, under-scoring successful trajectories may weaken the learning signal and slow policy~improvement.

Assessment reliability also depends on the VLM backbone, prompt formulation, and~API behavior. Prompt-paraphrase sensitivity and repeated-query stability were not evaluated, and~the experiments do not include a direct one-stage VLM success-judgment baseline. Table~\ref{tab:ablation_data_fusion} provides aggregate evidence for the value of action logs, but~paired qualitative comparisons with and without action logs are not~included.

The current implementation is suited to episodic or segment-level assessment rather than high-frequency control. Stage 1 accounts for more than 90\% of the measured latency. Moreover, the~controlled visibility setting simplifies visual interpretation, and~the evaluation is restricted to simulated ALFRED and AI2-THOR environments. Consequently, robustness to natural occlusion and real-world deployment remains unverified. Finally, all methods use replay buffers initialized with human-collected demonstrations; demonstration-free learning from scratch was not~evaluated.

\subsection{Future~Work}

Future work should evaluate the framework under natural occlusion, imperfect action execution, and~observations collected from physical robots. It should also include a direct one-stage VLM success-judgment baseline, test prompt-paraphrase and repeated-query stability, and~examine demonstration-free learning. Efficiency improvements should target Stage 1 through sparse frame selection, distilled or fine-tuned transition summarizers, and~reusable visual-context caches. Richer outputs could identify failed sub-steps or the observation segments that most influenced the final~score.

%%%%%%%%%%%%%%%%%%%%%%%%%%%%%%%%%%%%%%%%%%
\section{Conclusions} \label{sec:conclusions}

We presented \ours{}, an~action- and language-conditioned video assessment framework for trajectory-level task-progress feedback. The~framework summarizes visual transitions conditioned on executed actions and then evaluates the summary with respect to a natural-language instruction. When used as terminal feedback in the evaluated simulated household tasks, \ours{} achieved higher closed-loop task-completion rates than the tested static-image and embedding-similarity baselines and narrowed the gap to the ground-truth oracle. These findings are limited to the tested environments, VLM backbones, and~observation-based baselines. The~current implementation remains limited by conservative scoring, dependence on VLM selection and prompt formulation, controlled object visibility, and~substantial Stage 1~latency.

\vspace{6pt}

\authorcontributions{Conceptualization, H.K., J.J., and D.L.; methodology, H.K.; software, J.J.; validation, S.C. and H.Y.; formal analysis, D.L. and S.C.; investigation, D.L. and H.Y.; resources, C.D.Y.; data curation, S.C.; writing-original draft preparation, H.K. and J.J.; writing-review and editing, H.K., J.J., S.C., and H.Y.; visualization, S.C. and H.Y.; supervision, C.D.Y.; project administration, C.D.Y.; funding acquisition, C.D.Y. All authors have read and agreed to the published version of the~manuscript.}

\funding{This work was partly supported by Center for Applied Research in Artificial Intelligence (CARAI) grant funded by DAPA and ADD (UD230017TD), partly supported by Institute for Information \& communications Technology Planning \& Evaluation (IITP) grant funded by the Korea government (MSIT) (No.~RS-2021-II211381, Development of Causal AI through Video Understanding and Reinforcement Learning, and~Its Applications to Real Environments), and~partly supported by the National Research Foundation of Korea (NRF) grant funded by the Korea government (MSIT) (No.~RS-2025-24742969, Intelligent Robotic System using Continual Learning and Multimodal Language Model based Multi Attribute Feedback).}

% v1
% \funding{This work was partly supported by Center for Applied Research in Artificial Intelligence (CARAI) grant funded by DAPA and ADD (UD230017TD) and partly supported by Institute for Information \& communications Technology Planning \& Evaluation (IITP) grant funded by the Korea government(MSIT) (No.RS-2021-II211381, Development of Causal AI through Video Understanding and Reinforcement Learning, and~Its Applications to Real Environments).}

\institutionalreview{Not applicable.}

\informedconsent{Not applicable.}

\dataavailability{
The data used in this study are based on publicly available embodied instruction-following environments and benchmarks. The~ALFRED benchmark is available at \url{https://askforalfred.com} (accessed on 3 August 2026). Additional implementation details, prompts, and~experimental results are available from the corresponding author upon reasonable request.
}

\conflictsofinterest{
The authors declare no conflicts of~interest.
}

%%%%%%%%%%%%%%%%%%%%%%%%%%%%%%%%%%%%%%%%%%
%% Optional

%% Only for journal Encyclopedia
%\entrylink{The Link to this entry published on the encyclopedia platform.}

\abbreviations{Abbreviations}{
The following abbreviations are used in this manuscript:
\\

\noindent 
\begin{tabular}{@{}ll}
AI2-THOR & Allen Institute for Artificial Intelligence The House Of inteRactions\\
\ours{} & \fullours{}\\
CLIP & Contrastive Language-Image Pre-training\\
IQL & Implicit Q-Learning\\
R3M & Reusable Representation for Robotic Manipulation\\
RGB & Red-Green-Blue\\
VLM & Vision-Language Model
\end{tabular}
}

%%%%%%%%%%%%%%%%%%%%%%%%%%%%%%%%%%%%%%%%%%
%% Optional
\appendixtitles{yes} % Leave argument "no" if all appendix headings stay EMPTY (then no dot is printed after "Appendix A"). If~the appendix sections contain a heading then change the argument to "yes".
\appendixstart
\appendix
% \section[\appendixname~\thesection]{}
% \subsection[\appendixname~\thesubsection]{}
% The appendix is an optional section that can contain details and data supplemental to the main text---for example, explanations of experimental details that would disrupt the flow of the main text but nonetheless remain crucial to understanding and reproducing the research shown; figures of replicates for experiments of which representative data are shown in the main text can be added here if brief, or as Supplementary Data. Mathematical proofs of results not central to the paper can be added as an appendix.

% \begin{table}[H] 
% \caption{This is a table caption.\label{tab5}}
% %\newcolumntype{C}{>{\centering\arraybackslash}X}
% \begin{tabularx}{\textwidth}{CCC}
% \toprule
% \textbf{Title 1}	& \textbf{Title 2}	& \textbf{Title 3}\\
% \midrule
% Entry 1		& Data			& Data\\
% Entry 2		& Data			& Data\\
% \bottomrule
% \end{tabularx}
% \end{table}

% \section[\appendixname~\thesection]{}
% All appendix sections must be cited in the main text. In the appendices, Figures, Tables, etc. should be labeled, starting with ``A''---e.g., Figure A1, Figure A2, etc.

\appendix

\section{Implementation and Environment~Details}
\label{app:implementation_environment}
\unskip

\subsection{Downstream Policy Optimization~Details}
\label{app:policy_optimization}

Although the main focus of \ours{} is action- and language-conditioned video assessment and trajectory-level task assessment, we use an online off-policy control module to evaluate whether the generated feedback can support autonomous task execution. Specifically, we use IQL~\cite{kostrikov2022offline} as the downstream policy optimization algorithm, following prior work on reinforcement learning in ALFRED-like embodied environments~\cite{zhang2023bootstrap, shridhar2020alfred}. The~hyperparameters used in our experiments are summarized in Table~\ref{tab:iql_hyperparameters}.

\begin{table}[H]
\caption{Hyperparameters used for downstream off-policy policy~optimization.\label{tab:iql_hyperparameters}}
%\centering

\begin{tabularx}{1\textwidth}{lC}
\toprule
\textbf{Parameter} & \textbf{Value} \\
\midrule
Batch size & 128 \\
Training steps & 800k for Bedroom, 500k otherwise \\
Learning rate & $1 \times 10^{-4}$ \\
Optimizer & AdamW \\
Dropout rate & 0.1 \\
Weight decay & 0.1 \\
Discount factor $\gamma$ & 0.97 \\
Q update Polyak averaging coefficient & 0.005 \\
Policy and Q update period & 8 per training~iteration \\
Advantage clipping & $[0, 100]$ \\
IQL inverse temperature $\beta$ & 5 \\
IQL expectile parameter $\tau$ & 0.5 \\
Maximum context length & 8 \\
\bottomrule
\end{tabularx}
\end{table}

During training, the~buffer is initialized with a small number of human-collected demonstrations to reduce exploration time in the long-horizon embodied environments, following the initialization protocol used in prior ALFRED-based policy learning settings~\cite{zhang2023bootstrap}. These demonstrations are inserted into the replay buffer only; the policy is not pretrained through behavioral cloning or another imitation-learning objective. All compared methods use the same initialization protocol. The~experiments therefore do not evaluate demonstration-free learning from scratch. For~\ours{}, the~progress feedback scores generated by the vision-language evaluator are stored with the corresponding trajectories and used as trajectory-level feedback for downstream off-policy~optimization.

\subsection{Embodied Environment~Details}
\label{app:environment_details}

We evaluate \ours{} using a modified version of the ALFRED benchmark~\cite{shridhar2020alfred}, which provides visually grounded household environments and natural language instructions for embodied task execution. ALFRED is built on the AI2-THOR simulator~\cite{kolve2017ai2} and was originally designed for imitation learning. We use a modified version that supports reinforcement-learning-based interaction through a gym-style interface~\cite{zhang2023bootstrap}.

We further modify the environment to improve the visibility of manipulated objects. In~the original setup, when the agent picks up certain objects, the~object may be partially or fully occluded by the agent's view, especially for larger objects. Since \ours{} evaluates trajectories from visual observations, we ensure that picked-up objects remain clearly visible in the agent's view. This controlled visibility setting is applied to every evaluated method. It supports consistent comparison under observable object states, but~the present experiments do not evaluate robustness to natural occlusion or real-world visual~ambiguity.

We define evaluation tasks by randomly sampling 10 tasks from each of four unseen ALFRED floor plans, corresponding to Kitchen, Bedroom, Living Room, and~Bathroom environments. This results in 40 tasks in total. Each task is constrained to contain two sub-tasks. For~tasks originally containing more than two sub-tasks, only the first two sub-tasks are used. An~episode is considered successful only when both sub-tasks are~completed.

The agent receives RGB visual observations with a resolution of $224 \times 224$. For~downstream policy optimization and baseline methods, each observation is encoded using a frozen ResNet-18 pre-trained on ImageNet~\cite{he2016deep}, resulting in a $512 \times 7 \times 7$ visual feature representation. The~ALFRED action space consists of five navigation actions: \texttt{MoveAhead}, \texttt{RotateRight}, \texttt{RotateLeft}, \texttt{LookUp}, and~\texttt{LookDown}; and seven interaction actions: \texttt{Put}, \texttt{Pickup}, \texttt{Open}, \texttt{Close}, \texttt{ToggleOn}, \texttt{ToggleOff}, and~\texttt{Slice}~\cite{shridhar2020alfred}. For~interaction actions, the~policy additionally predicts 1 of 82 object types. Due to the large discrete action space, invalid action--object combinations are masked during policy optimization, following prior work~\cite{zhang2023bootstrap}. For~the \ours{} video interpretation prompt, we use the executed action names but do not include object-type~predictions.

Table~\ref{tab:stage_latency} reports the per-stage querying latency measured for each VLM backbone, complementing the latency--reliability trade-off shown in Figure~\ref{fig:pareto_latency_reliability}.

\begin{table}[H]
\caption{Per-stage querying latency for the benchmarked VLM backbones. Stage 1 (Action-Induced Transition Summarization) dominates the end-to-end cost across all backbones, whereas Stage 2 (Task-Progress Assessment) operates on compressed text and remains~negligible.\label{tab:stage_latency}}
%\centering
\tabcolsep=0.395cm
\begin{tabular}{lcccc}
\toprule
\textbf{VLM} & \textbf{Stage 1 (s)} & \textbf{Stage 2 (s)} & \textbf{Total (s)} & \textbf{Stage 1 Share (\%)} \\
\midrule
Gemini 1.5 Flash & 15.7  & 1.3 & 17.0  & 92.4 \\
GPT-4o mini      & 38.8  & 0.8 & 39.6  & 98.0 \\
Gemini 1.5 Pro   & 25.0  & 1.6 & 26.6  & 94.0 \\
Qwen2-VL-72B     & 912.2 & 3.6 & 915.8 & 99.6 \\
GPT-4o           & 29.5  & 0.8 & 30.3  & 97.4 \\
\bottomrule
\end{tabular}
\end{table}
\unskip

\section{Evaluation Task~Lists}
\label{app:task_details}

Tables~\ref{tab:kitchen_tasks}--\ref{tab:bathroom_tasks} list the task instructions used in the four embodied environments. These instructions are included to clarify the language-conditioned tasks used for trajectory-level evaluation and to support~reproducibility.

\begin{table}[H]
\caption{Task instructions used in the Kitchen~environment.\label{tab:kitchen_tasks}}
%\centering

\begin{adjustwidth}{-\extralength}{0cm}
%\centering %% If there is a figure in wide page, please release command \centering, for Table, ``\textwidth" should be ``\fulllength"
\begin{tabularx}{\fulllength}{c l >{\raggedright\arraybackslash}X}
\toprule
\textbf{Task No.} & \textbf{Sub-Task Type} & \textbf{Instruction} \\
\midrule
1 & PickupObject & Pick up the spoon from the~counter. \\
  & PutObject & Put the spoon in the white cup on the~shelf. \\
\midrule
2 & PickupObject & Pick up the egg that is beside the fork in the~sink. \\
  & CoolObject & Open the refrigerator, then place the egg on the glass shelf and close the fridge. Wait, then open the fridge and pick up the egg, then close the~fridge. \\
\midrule
3 & PickupObject & Pick up the tomato from the~sink. \\
  & CoolObject & Open the fridge door, put the tomato inside of the fridge, close the door, open the door, take the tomato out, close the~door. \\
\midrule
4 & PickupObject & Pick up the mug in the coffee~maker. \\
  & CoolObject & Open the fridge, put the cup in the fridge, close the fridge, wait, open the fridge, pick the cup, close the~fridge. \\
\midrule
5 & PickupObject & Pick up the~bread. \\
  & CoolObject & Open the fridge, put the bread in the fridge, close the fridge, open the fridge, get the bread, and~close the~fridge. \\
\midrule
6 & PickupObject & Pick up the white coffee cup to the right of the~trophy. \\
  & CleanObject & Put the coffee cup in the sink, turn on the water, turn off the water, and~pick up the coffee~cup. \\
\midrule
7 & PickupObject & Pick up the smaller silver knife on the~counter. \\
  & PutObject & Put the knife in the green cup in the~sink. \\
\midrule
8 & PickupObject & Pick up a bowl from the~shelf. \\
  & PutObject & Put the bowl on the~counter. \\
\midrule
9 & PickupObject & Grab the knife from the~counter. \\
  & PutObject & Put the knife in the pan on the~stove. \\
\midrule
10 & PickupObject & Pick up the knife from the~counter. \\
   & CleanObject & Place the knife in the sink and turn the water on. Turn the water off and pick up the~knife. \\
\bottomrule
\end{tabularx}
\end{adjustwidth}
\end{table}
\unskip

\begin{table}[H]
\caption{Task instructions used in the Bedroom~environment.\label{tab:bedroom_tasks}}
\centering

\begin{adjustwidth}{-\extralength}{0cm}
%\centering %% If there is a figure in wide page, please release command \centering, for Table, ``\textwidth" should be ``\fulllength"
\begin{tabularx}{\fulllength}{c l >{\raggedright\arraybackslash}X}
\toprule
\textbf{Task No.} & \textbf{Sub-Task Type} & \textbf{Instruction} \\
\midrule
1 & PickupObject & Pick up the bowl from the~shelf. \\
  & ToggleObject & Turn on the lamp sitting on the desk while holding the~bowl. \\
\midrule
2 & PickupObject & Pick up the white mug from the~desk. \\
  & ToggleObject & Turn the desk lamp on with the mug in~hand. \\
\midrule
3 & PickupObject & Pick up the book from the~bed. \\
  & ToggleObject & Turn on the lamp on the desk while carrying the~book. \\
\midrule
4 & PickupObject & Pick up the mug from the~shelf. \\
  & ToggleObject & Turn the lamp on while holding the~cup. \\
\midrule
5 & PickupObject & Pick up the bowl on the~desk. \\
  & ToggleObject & Turn on the lamp on the desk while holding the~bowl. \\
\midrule
6 & PickupObject & Pick up the pencil from the~desk. \\
  & PutObject & Put the pencil in the~bowl. \\
\midrule
7 & PickupObject & Pick up the alarm clock from the~desk. \\
  & ToggleObject & Turn on the lamp on the desk while holding the alarm~clock. \\
\midrule
8 & PickupObject & Pick up the clock from the back of the~desk. \\
  & ToggleObject & Hold the clock and turn on the lamp on the right side of the~desk. \\
\midrule
9 & PickupObject & Pick up the mug on the~shelf. \\
  & PutObject & Put the mug on the~desk. \\
\midrule
10 & PickupObject & Pick up the pencil on the~desk. \\
   & PutObject & Place the pencil in the glass bowl on the~desk. \\
\bottomrule
\end{tabularx}
\end{adjustwidth}
\end{table}
\unskip

\begin{table}[H]
\caption{Task instructions used in the Living Room~environment.\label{tab:livingroom_tasks}}

\begin{adjustwidth}{-\extralength}{0cm}
%\centering %% If there is a figure in wide page, please release command \centering, for Table, ``\textwidth" should be ``\fulllength"
\begin{tabularx}{\fulllength}{c l >{\raggedright\arraybackslash}X}
\toprule
\textbf{Task No.} & \textbf{Sub-Task Type} & \textbf{Instruction} \\
\midrule
1 & PickupObject & Pick up the cell phone from the~dresser. \\
  & ToggleObject & Hold the cell phone and turn the lamp~on. \\
\midrule
2 & PickupObject & Pick up the remote that is on the blue~chair. \\
  & ToggleObject & Turn on the lamp with the remote in~hand. \\
\midrule
3 & PickupObject & Pick up the laptop on the right after closing~it. \\
  & ToggleObject & Turn on the floor lamp while carrying the~laptop. \\
\midrule
4 & PickupObject & Pick the phone up from the~desk. \\
  & ToggleObject & Turn the lamp on while holding the~phone. \\
\midrule
5 & PickupObject & Grab the tissue paper from the~dresser. \\
  & ToggleObject & Carry the tissue as you turn on the~lamp. \\
\midrule
6 & PickupObject & Pick up the remote from the middle of the dresser, directly behind the~tissues. \\
  & ToggleObject & Hold the remote and turn on the~lamp. \\
\midrule
7 & PickupObject & Pick up a pillow from the~chair. \\
  & PutObject & Put the pillow on the~couch. \\
\midrule
8 & PickupObject & Pick up the statue on the top~shelf. \\
  & ToggleObject & Turn on the lamp while holding the~statue. \\
\midrule
9 & PickupObject & Pick up a statue from the~dresser. \\
  & ToggleObject & Turn on the floor lamp with the statue in~hand. \\
\midrule
10 & PickupObject & Pick up the left pillow on the~chair. \\
   & PutObject & Put the pillow on the sofa right of the~newspaper. \\
\bottomrule
\end{tabularx}
\end{adjustwidth}
\end{table}
\unskip

\begin{table}[H]
\caption{Task instructions used in the Bathroom~environment.\label{tab:bathroom_tasks}}

\begin{adjustwidth}{-\extralength}{0cm}
%\centering %% If there is a figure in wide page, please release command \centering, for Table, ``\textwidth" should be ``\fulllength"
\begin{tabularx}{\fulllength}{c l >{\raggedright\arraybackslash}X}
\toprule
\textbf{Task No.} & \textbf{Sub-Task Type} & \textbf{Instruction} \\
\midrule
1 & PickupObject & Pick up the bar of soap on the back of the~toilet. \\
  & PutObject & Place the soap in the trash~can. \\
\midrule
2 & PickupObject & Pick up bar of~soap. \\
  & CleanObject & Put soap in sink, turn water on, turn water off, remove soap from~sink. \\
\midrule
3 & PickupObject & Pick up the cloth from the~counter. \\
  & CleanObject & Put the cloth in the sink and turn the water on and then off and pick the cloth up from the~sink. \\
\midrule
4 & PickupObject & Pick the soap up from the back of the~toilet. \\
  & CleanObject & Put the soap in the sink and turn the water on and then off and pick up the soap~again. \\
\midrule
5 & PickupObject & Pick the cloth up from the~counter. \\
  & CleanObject & Put the cloth in the sink and turn the water on and then off and take the cloth out of the~sink. \\
\midrule
6 & PickupObject & Pick up the bar of~soap. \\
  & CleanObject & Put the bar of soap in the sink, turn the water on and then off and then pick up the bar of~soap. \\
\midrule
7 & PickupObject & Pick up the bar of soap on the back of the~toilet. \\
  & PutObject & Open the cabinet, put the bar of soap inside, and~close the~cabinet. \\
\midrule
8 & PickupObject & Grab a bar of soap off of the~counter. \\
  & PutObject & Put the soap in the trash~can. \\
\midrule
9 & PickupObject & Pick up the soap on the~counter. \\
  & PutObject & Open the cabinet and put in the soap then close the~cabinet. \\
\midrule
10 & PickupObject & Pick up toilet roll from off the~toilet. \\
   & PutObject & Open sink cabinet and place roll inside before closing the~door. \\
\bottomrule
\end{tabularx}
\end{adjustwidth}
\end{table}

%%%%%%%%%%%%%%%%%%%%%%%%%%%%%%%%%%%%%%%%%%
%\isPreprints{}{% This command is only used for ``preprints''.
\begin{adjustwidth}{-\extralength}{0cm}

\reftitle{References}

\PublishersNote{}
%\isPreprints{}{% This command is only used for ``preprints''.
\end{adjustwidth}
\end{document}